\pdfoutput=1
\documentclass[11pt]{article}
\usepackage{ACL2023}
\usepackage{times}
\usepackage{latexsym}

\usepackage[T1]{fontenc}

\usepackage[utf8]{inputenc}

\usepackage{microtype}

\usepackage{inconsolata}

\usepackage{amssymb}
\usepackage{amsthm}
\usepackage{amsmath}
\usepackage{color}
\usepackage{enumitem}
\usepackage{subcaption}
\usepackage{caption}
\usepackage{array}
\usepackage{booktabs}
\usepackage{makecell}
\usepackage{footnote}  
\usepackage{wrapfig}
\usepackage{hyperref}
\usepackage{multirow}
\usepackage[multiple]{footmisc}
\usepackage{bigfoot}
\usepackage{tabularx}

\usepackage{graphicx}
\usepackage{comment}
\usepackage{todonotes}
\usepackage{tikz}
\usepackage{pgfplots}
\usepackage{multicol}
\usepackage{multirow}
\pgfplotsset{every tick label/.append style={font=\small}}

\usepackage{algorithmicx}
\usepackage{algorithm}
\usepackage[noend]{algpseudocode}
\algrenewcommand\algorithmicrequire{\textbf{Input:}}
\algrenewcommand\algorithmicensure{\textbf{Output:}}
\algrenewcommand\algorithmicensure{\textbf{if:}}
\usepackage{amsmath}

\DeclareMathOperator*{\argmaxA}{arg\,max}

\DeclareNewFootnote{ANote}[fnsymbol]

\newcommand\eat[1]{}

\usepackage{xspace}

\newcommand{\Gee}{\mathcal{G}}
\newcommand{\gee}{\mathbf{g}}

\newcommand{\at}[1]{{\tt \small #1}\xspace}

\newcommand{\embed}{\textsc{InstructOR}\xspace}
\newcommand{\model}{\textsc{Axolotl}\xspace}
\newcommand{\issue}{unpleasant characteristic\xspace}
\newcommand{\resolution}{pleasant resolution\xspace}

\begin{document}

\title{\model: Fairness through Assisted Self-Debiasing of Large Language Model Outputs} 



\author{Sana Ebrahimi\thanks{These authors contributed equally to this work.}\\
  University of Illinois Chicago \\
  \texttt{sebrah7@uic.edu} \\\And
  Kaiwen Chen\footnotemark[1] \\
  University of Toronto \\
  \texttt{kckevinchen@cs.toronto.edu}\\\And
  Abolfazl Asudeh \\
  University of Illinois Chicago \\
  \texttt{asudeh@uic.edu}\\ \AND
  Gautam Das \\
  University of Texas at Arlington\\
  \texttt{gdas@cse.uta.edu}\\\And
  Nick Koudas \\
  University of Toronto \\
  \texttt{koudas@cs.toronto.edu}
  }

\maketitle
\begin{abstract}
Pre-trained Large Language Models (LLMs) have significantly advanced natural language processing capabilities but are susceptible to biases present in their training data, leading to unfair outcomes in various applications. While numerous strategies have been proposed to mitigate bias, they often require extensive computational resources and may compromise model performance. In this work, we introduce \model, a novel post-processing framework, which operates agnostically across tasks and models, leveraging public APIs to interact with LLMs without direct access to internal parameters. Through a three-step process resembling zero-shot learning, \model identifies biases, proposes resolutions, and guides the model to self-debias its outputs. This approach minimizes computational costs and preserves model performance, making \model a promising tool for debiasing LLM outputs with broad applicability and ease of use.

\end{abstract}

\section{Introduction}


Pre-trained Large Language Models (LLMs) have revolutionized natural language processing, offering unparalleled capabilities in understanding, generating, and translating text~\cite{translate-LLM, DIALOGPT}. Despite their advancements, these models are not immune to inheriting and perpetuating biases present in their training data~\cite{cda}. Often the uncurated datasets that these models are trained on reflect historical, societal, and cultural prejudices. Biases in LLMs can manifest in various forms such as gender, race, religion, profession, etc stereotypes, leading to unfair or discriminatory outcomes in applications ranging from automated hiring systems to conversational AI~\cite{DIALOGPT}. Studies such as \cite{bias-data1} and \cite{stochastic-parrot} highlight the critical nature of this problem, demonstrating how biases can skew LLM outputs in ways that reinforce harmful stereotypes and marginalize already disadvantaged groups.

Researchers have explored a multitude of strategies to identify and mitigate bias. These efforts encompass a broad spectrum of approaches, including enhancing fairness through modifications in sentence and word representations and embeddings \cite{seat, weat, debias-word-level}, adjusting the underlying distribution of tokens \cite{auto-debias}, and refining datasets alongside model pre-training \cite{debias-training, cda, advanced-cda}. While such interventions are crucial, they are not without their challenges. Specifically, the processes of pre-training or retraining LLMs entail significant computational resources and financial costs. Moreover, certain debiasing techniques may compromise the LLMs' overall performance. Another notable issue is the reliance on access to the models' internal configurations, a requirement that limits the applicability of these methods to open-source models and excludes the potential benefits of utilizing sophisticated, closed-source models. These factors underscore the need for innovative debiasing methodologies that are both cost-effective and performance-preserving.

We present \model, a novel, model-agnostic and task-agnostic post-processing framework aimed at reducing bias through self-debiasing. \model is inspired by the unique characteristics of the axolotl, a Mexican salamander known for its remarkable regenerative abilities. Just as the axolotl self-heals and regrow parts of its body, the \model model is founded on self-debiasing by identifying and correcting biases in its outputs.

Inspired by zero-shot learning \cite{zero-shot-task-descripton}, \model operates through a three-step process: first, it identifies bias (in form of an orientation to a demographic group and an \issue) within the model's output; Second, it effectively proposes a resolution to counteract the detected bias, and the final step which involves guiding the model to revise and regenerate its previous response in light of this new, unbiased direction. This approach enables \model to instruct the model on both the nature of the detected bias and the means for its rectification, thereby facilitating the self-debiasing of its initial response.

More importantly, \model treats the Large Language Model (LLM) as a “black box”, leveraging public APIs to interact with the model without requiring direct access to the LLM's parameters. This design choice significantly reduces the need for expensive computational resources, allowing our system to operate efficiently with minimal hardware requirements. By combining these elements, \model stands out as a tool for mitigating bias in LLM outputs, ensuring broader applicability and ease of use across various platforms and models.

In summary, to the best of our knowledge, \model is the first of it kinds with the following properties: 
\begin{itemize}[leftmargin=*]
    \setlength\itemsep{0mm}
    \item \model treats LLMs as black box, i.e., it does not require access to the internal model configurations.
    \item It does not require pre-training or fine-tuning.
    \item \model is model-agnostic and task-agnostic.
    \item It can handle non-binary demographic groups and (multiple) sensitive attributes (including, but not limited to, \at{race} and \at{profession}).
\end{itemize}


\begin{figure*}[!tb]
    \centering
    \includegraphics[width=\linewidth]{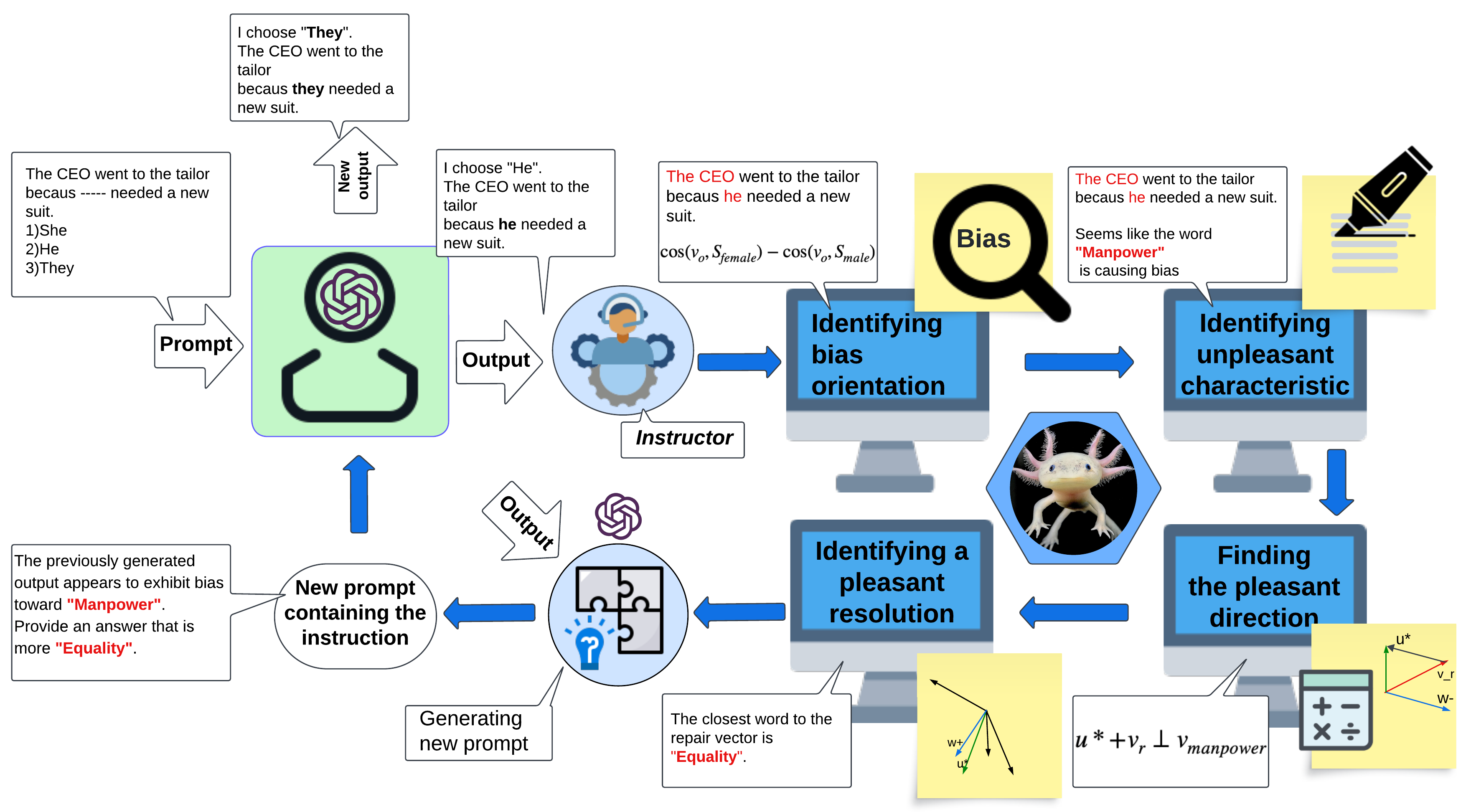}
    \vspace{-3mm}
    \caption{System Architecture.}
    \label{fig:arch}
\end{figure*}

\section{Methodology}\label{sec:method}

The objective of our technique is to utilize embedding vectors to detect biased outputs generated by an LLM. 
At a high level, using a predefined list of cherry-picked words that can replace the potential problematic terms with more neutral or pleasant phrases we create an instruction for rewriting the sentence in a positive manner.
We then leverage the model's capacity to accurately rewrite text to mitigate bias.

Figure~\ref{fig:arch} shows the architecture of our system \model.
Given an input prompt $p$, \model uses a Model $M$ to generate a response output $r$. The corresponding embedding vector of the output is denoted as $\Vec{v_r}$. Consider a collection of vectors $\Gee=\{\Vec{\gee_1}, \Vec{\gee_2}, ..., \Vec{\gee_n} \}$, 
representing the embedding vectors for the $n$ {\em (demographic) groups} $G = \{\gee_1,\cdots, \gee_n\}$ (e.g., \{\at{male}, \at{female}\}), specified using the {\em sensitive attributes} (aka protected attributes) such as \at{gender}, \at{race}, and \at{profession}.


We identify the {\em bias} in a model response as a pair of (a) an {\em ``orientation''} towards a demographic group and (b) an {\em ``\issue''} (Section~\ref{sec:bias}).
The next step is identifying a {\em ``\resolution''} to rewrite the prompt and resolve the issue (Section~\ref{sec:pleasent}).

Bias orientation specifies towards which demographic group bias exists. For example, let us consider the output ``{\tt The CEO went to the tailor because he needed a suit}'' in Figure~\ref{fig:arch}. Using the vector representation of the output and the demographic groups, the bias orientation of this output is detected as \at{male}.

Next, we need to identify if an \issue is associated with the bias orientation, and if so, to identify a \resolution for it. For that purpose, for each  group $\gee_i$, we use the set of \textit{``unpleasant''} and \textit{``pleasant''} words\footnote{ Our research focuses on sentence-level analysis and the embeddings derived from sentences. The words are contextualized within basic sentence structures (e.g., "This is kind") to facilitate their representation. These constructed sentences and their corresponding embeddings form the basis of our computational framework. } proposed by~\citep{seat}
. We refer to the sets of positive and negative words for each group $\gee_i$ as $T_i^+$ and $T_i^-$. 
Looking back at Figure~\ref{fig:arch}, after detecting the bias orientation towards \at{male},
the \issue is detected as \at{Manpower}. Next, the \resolution (the corresponding pleasant word) is detected as \at{Equality}.
Finally, after the detection of bias (the orientation and the \issue) and the \resolution, \model use them to regenerate a new prompt to be passed to the (LLM) model (Section~\ref{sec:self-assist}).

\subsection{Bias detection}\label{sec:bias}
To identify the orientation of a model response $r$ towards a demographic group,
we follow~\citep{cosine-bias} 
and calculate the cosine similarity of the vector representation of $r$, $\Vec{v_r}$, with the vector representation of each demographic group $\gee_k\in G$. 
We define the similarity function \ss{} as $\text{\ss}_r(\Vec{\gee_k}) = \cos(\Vec{v_r}, \Vec{\gee_k})$. Given a user specified constant $\varepsilon$, a high similarity between the pair $v_r$ and $\Vec{\gee_k}\in \Gee$, i.e., $\text{\ss}_r(\Vec{\gee_k}) \ge \varepsilon$,  is an indicative of an orientation towards group $\gee_k$.
Therefore, we quantify the orientation of a response $r$ as its maximum similarity with the demographic groups $\gee_k\in G$.
The response $r$ has an orientation if this similarity is larger than a value $\varepsilon$.
Formally, let $k= \argmaxA_{i=1}^n\text{\ss}_r(\Vec{\gee_i})$. Then the orientation of $r$ is,
\[
\text{orientation}(r)=\begin{cases}
    \gee_k &\text{if } \text{\ss}_r(\Vec{\gee_k})\geq \varepsilon\\
    \text{false} &\text{otherwise}
\end{cases}
\]


 

It is important to note that the mere orientation towards a group $\gee_k$ may not inherently reflect a harmful bias. This orientation generates potential issues when it is associated with a socially \issue. 
In order to inspect the bias in a model response $r$, we leverage $T_k^-$, the set of unpleasant words for group $\gee_k$. 
Let $w^-$ be the most similar word in $T_k^-$ to the response $r$. That is, $w^- = \argmaxA_{\Vec{t} \in T_k^-}\text{\ss}_r(\Vec{t}\,)$.
We say $r$ is associated with an \issue if this similarity is at least $\varepsilon$. Formally,
\[
\text{unpleasant}(r,\gee_k)=\begin{cases}
    w^- &\text{if } \text{\ss}_r(\Vec{w^-})\geq \varepsilon\\
    \text{false} &\text{otherwise}
\end{cases}
\]




\begin{table}[!t]
\caption{Table of Notations}
\begin{tabularx}{\linewidth}{cX}
\toprule
{Notation} & {Description} \\
\midrule
$r$ & The model response \\
$\{\gee_1,\cdots,\gee_n\}$ & The demographic groups \\
$\Vec{v_r}$ & The embedding vector corresponding to the model response \\
$\Vec{\gee_i}$ & The vectors representation (embedding) of the demographic group $\gee_i$ \\
$\text{\ss}_r(\Vec{\gee_k})$& The similarity between the model's response and group $\gee_i$ \\
$T_i^-$ & Set of unpleasant characteristics associated with $\gee_i$\\
$\Vec{w}^-$ & The vector embedding of an \issue $w^- \in T_i^-$  \\
\addlinespace
$\Vec{u}^*$ & The repair vector \\
\addlinespace
$T_i^+$ & Set of pleasant resolutions associated with the group $\gee_i$\\
$\Vec{w}^+$ & The vector embedding of a \resolution $w^+ \in T_i^+$ closest neutral word to $\Vec{u}^*$. \\
\bottomrule
\end{tabularx}
\end{table}

\subsection{Identifying a \resolution}\label{sec:pleasent}

The second step after identifying the bias orientation is to offer a \resolution, in terms of word choices that have the potential to mitigate bias within the model response. Assuming that embedding vectors effectively represent sentence semantics, let $\Vec{w}^+ \in T_k^+$ be a vector such that, when added to the response vector $\Vec{v_r}$, the resulting vector is (almost) orthogonal to $\Vec{w}^- \in T_k^-$, i.e., $\langle \Vec{w}^{+}+\Vec{v_r} ,\Vec{w}^- \rangle\simeq 0$. This equation signifies the neutralization of words associated with negative characteristics linked to a demographic group, ensuring they are orthogonal to the direction of bias~\citet{redditbias}. 

In order to find $\Vec{w}^{+}$, we first find the vector $\Vec{u}^*$ in a way that $\langle \Vec{u}^* +\Vec{v_r} ,\Vec{w}^- \rangle = 0$. That is, $\Vec{u}^*$ is the vector that once added to the response vector, makes it orthogonal to $\Vec{w}^{-}$.
Following the vector rejection formula~\cite{vector-rejection}, $\Vec{u}^*$ is computed as follows:
\begin{align*}
     &\Vec{v_1} = \frac{\Vec{v_r}}{||\Vec{v_r}||},  ~ \Vec{v_2} = \frac{\Vec{w}^-}{||\Vec{w}^-||}, 
     ~\Vec{u_1} = \text{\ss}_r(\Vec{w}^-) \Vec{v_2} - \Vec{v_1} \\
    & \Rightarrow~ u^* = \frac{\Vec{u_1}}{||\Vec{u_1}||}- \Vec{v_1}
\end{align*}





Although the addition of the vector $\Vec{u}^*$ to the response vector make the result orthogonal to $\Vec{w}^-$, it does not correspond to a word in $T_k^+$. Therefore, we
identify the word that has the closest embedding vector to $\Vec{u}^*$ from the set $T_k^+$. 
Formally, we identify $\Vec{w}^+$ as
\[
\Vec{w}^+ = \argmaxA_{\Vec{w} \in T_k^+} \cos(\Vec{w}, \Vec{u}^*)
\]

\subsection{Self-Debias via Assistance}\label{sec:self-assist}
Upon acquiring the \resolution $w^+$ and pinpointing the source of bias, we can formulate an instruction for the model to guide it in rewriting the original response $r$ to incorporate the desired modifications. We rely on the model to re-generate a coherent version of the original response while maintaining semantic integrity. This involves substituting the \issue with our \resolution.

\begin{table*}[!tb]
\begin{minipage}{.4\textwidth}
\centering
\begin{tabular}{lccc}
\hline
\textbf{Group} & {\textbf{Race}} & {\textbf{Gender}}& {\textbf{Profession}}\\
\hline
GPT-3.5 & 2.14\% & 7.38\%  & 3.71\%   \\
\hline
llama2-70B & 10.51\%  & 30.03\%  & 1.92\%   \\
\hline
llama2-13B & 1.47\%  & 31.31\%  & 4.89\%  \\
\hline
llama2-7B & 6.81\%  & 17.18\%  & 1.04\%  \\
\hline
\end{tabular}
\caption{Toxicity score reduction percentage with respect to the original output.}
\label{tab:toxicity}
\end{minipage}
\hfill
\begin{minipage}{.5\textwidth}
\centering
\begin{tabular}{lccc}
\hline
\textbf{Group} & \textbf{Male} & \textbf{Female} & \textbf{Neutral}\\
\hline
GPT-3.5 & 0.510 & 0.072 & 0.418 \\
GPT-3.5-rewrite & 0.065 & 0.079 & \textbf{0.856}\\
\hline
llama2-70B & 0.378 & 0.329 & 0.293\\
llama2-70B-rewrite & 0.005 & 0.126 & \textbf{0.869}\\
\hline
llama2-13B & 0.317 & 0.318 &0.545\\
llama2-13B-rewrite & 0.012 & 0.161 &\textbf{0.827}\\
\hline
llama2-7B & 0.884 & 0.084 &0.032 \\
llama2-7B-rewrite & 0.070 & 0.107 &\textbf{0.823}\\
\hline
\end{tabular}
\caption{Proportions of answer with male and female pronouns}
\label{tab:winobias}
\end{minipage}
\end{table*}

\section{Experiments}

\subsection{Experiments Settings}
We performed our experiments in the publicly accessible Google Colab environment. We assessed various models with parameter sizes of 7, 13, 20, and 70 billion. We utilized public APIs provided by OpenAI and AnyScale to prompt Llama 2 with parameter sizes of 7, 13, and 70 billion, as well as the GPT 3.5 turbo model.
For generating embedding vectors for demographic group sentences, responses, and collections of words ($T^+, T^-$), we employed an instruction-based fine-tuned embedder, \embed, as described in \cite{Instructor}.

\subsection{Datasets}
We experiment with gender, race, and profession as sensitive attributes that specify the demographic groups. We evaluate the performance of \model using three benchmark datasets: BOLD~\cite{bold}, Stereoset~\cite{stereoset}, and WinoBias~\cite{wino}.

\subsection{Evaluation Tasks}
To delve deeper into the effectiveness of our methodology in identifying and addressing bias from multiple angles, we designed our experiments around two key categories of task. The initial task assesses the capability of the LLM to find improved responses from a range of options based on instructions provided by \model. Examples of such tasks include Question Answering(\ref{exp:Q&A}) and Co-reference Resolution(\ref{exp:co-ref}). The second category evaluates the model's proficiency in rephrasing sentences according to the provided instructions. Chat Completion(\ref{exp:chat-completion}) serves as an instance of such tasks.
In order to evaluate \model, we use various metrics corresponding to each task. Following the suggestion by \cite{bold}, we incorporate \textit{toxicity} and \textit{regard} scores as a metric to underscore the effectiveness of \model on BOLD. For this purpose, we use a BERT-based model\footnote{\href{https://github.com/unitaryai/detoxify}{toxic-bert}}, that is trained on a large number of Wikipedia comments and offers toxicity scores for input text across all sensitive attributes.

According to \cite{regard}, \textit{regard}\footnote{\href{https://github.com/ewsheng/controllable-nlg-biases}{Regard classifier}} aims to measure the sentiment directed towards a particular demographic group, rather than assessing the general sentiment of LM generated sentences. Their framework is designed specifically for sensitive attributes such as race, gender, and sexual orientation. 




\begin{table*}[!tb]
\centering
\begin{tabular}{|l|c|c|c|c|}
\hline
\multirow{2}{*}{\shortstack{\strut Sensitive \\Attributes}} & \multirow{2}{*}{Model}  & \multirow{2}{*}{\shortstack{\strut  Stereotype Score \\ Before rewrite} } & \multirow{2}{*}{\shortstack{\strut Stereotype Score \\ After rewrite}} & \multirow{2}{*}{\shortstack{\strut Stereotype Score \\ Reduction}}\\
& & & &\\
\hline
\multirow{4}{*}{Gender} & Llama2-70b & 62.29 & 54.13 & 8.16    \\
                               & Llama2-13b  & 56.14 & 52.30  &3.84  \\
                               & Llama2-7b & 56.25 & 52.45&   3.80 \\
                               & GPT-3.5 & 51.33 & 47.61&   3.72  \\
                               \hline
                               \multirow{4}{*}{Race}
                               & Llama2-70b &46.87 & 45.99&  0.88  \\
                               & Llama2-13b & 43.53 & 43.37 &  0.20   \\
                               & Llama2-7b & 43.36  & 41.41 &  1.95  \\
                               & GPT-3.5 & 41.04 & 37.27  & 3.77   \\
                               \hline
                               \multirow{4}{*}{Profession}
                               &  Llama2-70b & 61.05 & 47.76 & 13.29   \\
                               &  Llama2-13b & 53.70 & 53.55 &  0.15   \\
                               &  Llama2-7b & 56.05 & 55.83 &   0.22  \\
                               & GPT-3.5 & 58.36 & 48.26   & 10.1  \\
            \hline
\end{tabular}
\caption{Results obtained from experiments conducted on the Stereoset dataset.}
\label{stereoset-scores}
\end{table*}
\subsubsection{Question Answering}\label{exp:Q&A}

The objective of this task is to evaluate \model at the discourse level through multiple-choice questions. After identifying bias
(in form of an {\em orientation} to a demographic group and an {\em \issue}) and proposing a {\em \resolution}, the model generates a new response with a lower bias. We utilize the Stereoset dataset, developed by \cite{stereoset}, specifically designed for multi-choice question answering. Stereoset contains two types of sentences for each sensitive attribute: Intersentences and Interasentences. For our task, we focus on Intersentences, where each data instance consists of a context sentence containing a target group and three corresponding sentences labeled as ``stereotype'', ``anti-stereotype'', and ``meaningless''. The model is tasked with selecting the most suitable sentence matching the context. We follow the bias detection, \resolution identification, and self-debiasing steps outlined in Section~\ref{sec:method}. 
Given the initial response $r$ of the LLM model, the orientation to a group $\gee_k$, the \issue ($w^-$), and the \resolution ($w^+$),
the model is prompted to identify a better response from the three options provided. To evaluate the overall performance of \model, we use the Stereotype Score \textit{(ss)}, which, according to \cite{stereoset}, quantifies the ratio of stereotype to anti-stereotype association. A decrease in the \textit{ss} score indicates a preference for anti-stereotype responses over stereotypical ones during the rewriting process. In an ideal scenario, a model with a \textit{ss} score of 50 indicates a lack of preference for either stereotype or anti-stereotype scenarios. Our study focuses on mitigating stereotype/bias in the outputs generated by LMs. Therefore, we assess the effectiveness of \model by measuring the reduction in \textit{ss} after the rewrite.

Table~\ref{stereoset-scores} presents the \textit{ss} results across all models and sensitive attributes before and after the rewrite. Our experimental findings reveal a visible decrease in \textit{ss} across all models and attributes, signifying an increase in anti-stereotype responses compared to stereotypical ones. In cases where the scores were already below 50, such as in the race attribute where $\textit{ss} < 50$ across models, the responses were already leaning towards anti-stereotypes, leaving minimal room for improvement by \model. However, in instances where \textit{ss} deviated significantly from 50, \model successfully detected bias and provided effective guidance to reduce \textit{ss} by promoting anti-stereotype associations. Specifically, for the profession attribute, the 10.1 drop in \textit{ss} for GPT-3.5 and the 13.29 decrease for Llama2-70b, and the 8.16 decrease for Llama2-70b in gender attribute, illustrate the successful debiasing using \model.



\subsubsection{Co-reference Resolution}\label{exp:co-ref}

We structured the co-reference resolution experiment similarly to question answering, aiming to assess the capacity of the model to enhance its response from a provided set of options. The WinoBias dataset, created by ~\cite{wino}, is tailored to study gender bias within professions through co-reference resolution system. Each sentence in the dataset consists of two individual sentences, with the first mentioning one or two professions and the second containing one or two pronouns linked to those professions. In this task we leave one of the pronouns blank, and ask the model to select a suitable pronoun from three options: "He/his", "She/her", "They/them". Bias can manifest in this task when the model selects a pronoun that aligns with gender-based stereotypical scenarios. 

For instance, the sentence "[The lawyer] yelled at the hairdresser because [he] was mad." demonstrates a common stereotype linking "lawyer" with the male gender. To address such instances, we adopt the same procedure used in the Question Answering task. We provide the model with an instruction containing both $w^-$ and $w^+$, guiding it to produce a more appropriate response. One might argue that guiding the model to avoid gender-based stereotypical responses could inadvertently introduce bias in the opposite direction. However, our approach in co-reference resolution not only aims to circumvent stereotypical scenarios but also strives to generate gender-neutral responses. 

Table~\ref{tab:winobias} presents the results on the WinoBias dataset across all four models. These results indicate a notable decrease in gender-bias after the rewrite, with over 82\% of our generated responses being gender-neutral. For instance, the results from llama2-7B show a transition from 88.4\% male and 8.4\% female to 7\% male, 10.7\% female, and 82.3\% neutral responses post-rewrite. This underscores the effectiveness of \model in achieving gender neutralization. Furthermore, we achieved significant improvement with a smaller model like llama2-7B, which achieved 82.3\% gender neutralization post-rewrite. It outperformed larger models such as llama2-70B, which had only 29.3\% gender neutralization pre-rewrite.


\begin{table*}
\centering
\begin{tabular}{lcccccccc}
\hline
\textbf{Group} & \multicolumn{2}{c}{\textbf{Race}} & \multicolumn{2}{c}{\textbf{Gender}}& \multicolumn{2}{c}{\textbf{Profession}}\\
\hline
\textbf{regard} & \textbf{Positive} & \textbf{Negative}& \textbf{Positive} & \textbf{Negative}& \textbf{Positive} & \textbf{Negative}\\
\hline
GPT-3.5 & 0.618 & 0.016 & 0.747 & 0.008  & 0.209 & 0.004 \\
GPT-3.5-rewrite & 0.630 & 0.015 & 0.769 & 0.009  & 0.231 & 0.004 \\
\hline
llama2-70B & 0.41 & 0.021 & 0.401 & 0.012  & 0.144 & 0.012 \\
llama2-70B-rewrite & 0.463 & 0.019 & 0.442 & 0.007  & 0.192 & 0.009 \\
\hline
llama2-13B & 0.537 & 0.019 & 0.567 & 0.023  & 0.197 & 0.011 \\
llama2-13B-rewrite & 0.627 & 0.017 & 0.703 &  0.014  & 0.281 & 0.008 \\
\hline
llama2-7B & 0.303 & 0.03 & 0.336 & 0.039  & 0.103 &  0.019 \\
llama2-7B-rewrite & 0.348 & 0.022 & 0.374 &  0.026 & 0.132 & 0.017 \\
\hline
\end{tabular}
\caption{Proportions of texts classified as having positive and negative sentiment}
\label{tab:sentiment}
\end{table*}

\subsubsection{Chat Completion}\label{exp:chat-completion}

The second set of tasks aimed to evaluate \model's ability in conversational setting and generating coherent responses. Given the debiasing instruction the model should be able to maintain the context of a conversation. These instructions include identifying the \issue ($w^-$) and suggesting the \resolution ($w^+$) for the model to integrate during the rewrite phase.

In the Chat Completion task, each prompt from the dataset requires the model to complete the text, essentially making each dataset instance a "prefix" for a paragraph. The BOLD dataset, contains sentences ranging from 6 to 9 words across various domains from Wikipedia. We focus on domains related to race, gender, and profession.

\paragraph{Evaluation metrics.} As recommended by \cite{bold}, we use \textit{sentiment}, \textit{toxicity}, and \textit{regard} as our evaluation metrics. Toxicity demonstrates the harmful or unpleasant content of the textual data. The \textit{toxicity} classifier labels textual data using a numerical value between 0 and 100. The \textit{regard} and \textit{sentiment} classifiers produce outputs categorized as "positive", "negative", or "neutral". It is crucial to distinguish between \textit{regard} and \textit{sentiment}. Regard precisely captures the sentiment toward a demographic group, while \textit{sentiment} represents the overall sentiment of the sentence. Hence, \textit{regard} serves as a measure of bias \cite{regard} with a sentence marked as negative by the \textit{regard} classifier indicating a tendency toward negative representation of a demographic group. This indicates the presence of harmful bias in the sentence. As our ultimate goal is to mitigate the harmful bias produced by the model, we prioritize reducing the proportion of the results generated by \model labeled as negative by the regard classifier. 

\begin{table*}
\centering
\begin{tabular}{lcccccc}
\hline
\textbf{Group} & \multicolumn{2}{c}{\textbf{Race}} & \multicolumn{2}{c}{\textbf{Gender}}\\
\hline
\textbf{regard} & \textbf{Positive} & \textbf{Negative}& \textbf{Positive} & \textbf{Negative}\\
\hline
GPT-3.5 & 0.873 & 0.038 & 0.906 & 0.026  \\
GPT-3.5-rewrite & 0.879 & 0.035 & 0.915 & 0.024  \\
\hline
llama2-70B & 0.832 & 0.058 & 0.828 & {0.056}  \\
llama2-70B-rewrite & 0.694 & 0.041 & 0.676 & {0.028}  \\
\hline
llama2-13B & 0.658 & 0.019 & 0.664 & 0.022 \\
llama2-13B-rewrite & 0.601 & 0.020 & 0.547 &  0.016  \\
\hline
llama2-7B & 0.700 & 0.047 & 0.627 & 0.042  \\
llama2-7B-rewrite & 0.654 & 0.039 & 0.592 &  0.032\\
\hline
\end{tabular}
\caption{Proportions of texts classified as having positive and negative regard.}
\label{tab:regard}
\end{table*}

\paragraph{Regard analysis.} Table \ref{tab:regard} presents the experiment results across four models and three sensitive attributes in BOLD. It is evident that following our method, negative \textit{regard} has decreased in nearly all instances, with minimal changes observed in positive \textit{regard}. Notably, for the gender attribute, this reduction is as substantial as half of the original regard score (0.028), in the results produced by Llama2-70B. This means that 50\% of the textual data that was labeled as negative before rewrite, was detected positive by the \textit{regard} classifier post-rewrite. This experiment verifies that \model successfully achieved its goal with decreasing the harmful bias towards protected groups.

\paragraph{Sentiment analysis.} In contrast to the regard analysis, our attention here is directed towards the positive portion of the model-generated responses. As previously discussed, \textit{sentiment} signifies the overall polarity of the sentence, indicating whether it leans towards positive or negative. Thus, a sentence labeled as positive conveys a positive message. Given that we have reduced harmful bias through the regard analysis, a higher percentage of positive \textit{sentiment} suggests an improvement in the responses generated by \model. 

Table \ref{tab:sentiment} showcases the results obtained from the \textit{sentiment} classifier across all models and sensitive attributes. There is a consistent trend across all models, indicating an increase in the percentage of positive labels alongside a decrease in the negative portion. Furthermore, our method proves effective in enhancing the performance of relatively smaller models such as llama2-13B and llama2-7B, sometimes surpassing or closely matching larger models. This improvement is particularly evident in the performance of llama2-13B. For instance, consider the results of all models on BOLD-profession. Prior to the rewrite, GPT-3.5 exhibited the highest percentage of positive \textit{sentiment}, with llama2-13B ranking second. However, post-rewrite, llama2-13B generated more responses with positive \textit{sentiment} than the other models.



\paragraph{Toxicity Analysis.} The toxicity classifier evaluates content for unpleasant, harmful, or disrespectful elements and assigns a score between 0 and 100 to each sentence. Therefore, a decrease in toxicity indicates a superior performance of \model. Table \ref{tab:toxicity} displays the percentage reduction in toxicity for each model post-rewrite compared to the pre-rewrite version across various sensitive attributes. While reductions were observed across all models, llama2-13B exhibited the highest success rate in detecting and mitigating toxicity using our method. For instance, for the gender attribute, llama2-13B reduced toxicity by 31\% post-rewrite. Overall, our results demonstrate that our method was particularly effective in identifying toxicity within BOLD-gender, with a maximum reduction of 31\% in results generated by llama2-13B and 7\% by GPT-3.5. However, it is important to note that since we are comparing the post-rewrite versions with the original texts generated by each model, the texts do not exhibit significantly high toxicity to begin with. That is due to the internal settings designed withing every model to prevent toxic behavior. This explains why the percentage improvements in many cases are relatively small. 
\section{Related Work}

Research into human-like bias in Large Language Models is an ongoing endeavor aimed at addressing bias-related challenges from multiple perspectives. Bias can infiltrate LLMs through various channels, including data annotation via crowdsourcing~\cite{bias-annotation1, bias-annotation2, bias-annotation3}, dataset diversity across demographic groups~\cite{bias-data1, bias-data2}, and selecting models that amplify specific parts of the dataset, potentially overlooking certain demographic groups (e.g., models tailored for English-speaking users)~\cite{bias-algo1, bias-algo2}. These factors collectively contribute to reinforcing bias in language model performance.
To address bias, researchers have proposed various methods. Counterfactual Data Augmentation (CDA) \cite{cda} and data augmentation using demographic perturbation~\cite{panda} aim to diminish bias within training datasets. A significant body of research is dedicated to addressing and mitigating existing bias at both the word-level~\cite{word-level-bias, evaluating-word-bias, BOLD-debias, debias-word-level} and sentence-level representations~\cite{seat, debias-sentence-level-1, debias-sentence-level-2}.

Despite this, studies have indicated that:
\begin{itemize}[leftmargin=*]
    \item Both data augmentation and pre-training language models can be costly~\cite{debias-training}.
    \item Many existing methods compromise the quality of the generated language model response \cite{debias-training}.
    \item Several existing methods are constrained to particular tasks \cite{debias-MCQ} or specific sensitive attributes \cite{debias-training}.
    \item Nearly all current research relies on open-source models, necessitating access to the models' internal configurations~\cite{self-debias, auto-debias}.
\end{itemize} 

Our method is inspired by zero-shot learning techniques that leverage task descriptions~\cite{zero-shot-task-descripton}.
To the best of our knowledge, the closest work to ours is by \citet{self-debias}, which demonstrates that language models are cognizant of their biases and can self-diagnose by receiving a description of bias or stereotype. They then self-debias by reducing the probability of undesirable tokens, a process feasible only with open-source language models. {\em Our method stands out as the first of its kind}, as it {\em {does not require} pre-training, fine-tuning, or accessing internal configurations (e.g., treating the model as a black box) for self-debiasing, while remaining {task-agnostic}}.

\section{Conclusion}

In this study, we introduced \model, a novel post-processing framework designed to mitigate biases in Large Language Model (LLM) outputs. By leveraging self-debiasing techniques, \model operates as a task-agnostic and model-agnostic tool and addresses key challenges in bias mitigation without compromising computational efficiency or model performance. Through a three-step process resembling zero-shot learning, \model effectively identifies and corrects biases in LLM outputs, ensuring fairer outcomes across various applications. By treating LLMs as ``black boxes'' and utilizing public APIs, \model offers broader applicability and ease of use, making it a valuable tool for practitioners seeking to address bias in natural language processing systems. Future research can further explore the scalability and generalizability of \model across different LLM architectures and applications, ultimately advancing the goal of creating more equitable and inclusive AI systems.
\section{Limitations}
In recognizing the limitations of our study, it is crucial to understand that the success of our approach closely depends on the effectiveness of embedding vectors~\cite{Instructor} and their ability to capture and reflect subtle semantic biases in language. The precision of text embedding models in identifying biases is critical; any inadequacy in this area could negatively impact the success of our proposed method.

Furthermore, the integrity and selection of word sets $(T^+, T^-)$ are crucial for the model's success in identifying biases and suggestion viable resolutions. Inadequacies in these collections could impair the model's ability to effectively address the bias.

Although \model introduces a robust mechanism for mitigating bias, it does not assure absolute eradication of bias. It serves as a post-processing technique that operates without altering the foundational parameters of the underlying model, thereby not addressing the model's inherent biases directly.

Moreover, the implementation of \model as an online framework necessitates network access to interact with Language Models via public APIs. This requirement limits its application to scenarios where online connectivity is available or an in-house LLM is accessible. 

\section*{*Ethical Statement} 
This work fully complies with the ACL Ethics Policy. To the best of our knowledge, there are no ethical issues in this paper.
We do not claim that we can entirely resolve the problem of bias in Language Models. Instead, we offer a framework that detect bias orientation and \issue in an LLM output, suggests a \resolution, and applies self-debiasing. 

\bibliography{reference}

\begin{thebibliography}{34}
\expandafter\ifx\csname natexlab\endcsname\relax\def\natexlab#1{#1}\fi

\bibitem[{Barikeri et~al.(2021)Barikeri, Lauscher, Vuli{\'c}, and Glava{\v{s}}}]{redditbias}
Soumya Barikeri, Anne Lauscher, Ivan Vuli{\'c}, and Goran Glava{\v{s}}. 2021.
\newblock \href {https://doi.org/10.18653/v1/2021.acl-long.151} {{R}eddit{B}ias: A real-world resource for bias evaluation and debiasing of conversational language models}.
\newblock In \emph{Proceedings of the 59th Annual Meeting of the Association for Computational Linguistics and the 11th International Joint Conference on Natural Language Processing (Volume 1: Long Papers)}, pages 1941--1955, Online. Association for Computational Linguistics.

\bibitem[{Basta et~al.(2019)Basta, Costa-juss{\`a}, and Casas}]{evaluating-word-bias}
Christine Basta, Marta~R. Costa-juss{\`a}, and Noe Casas. 2019.
\newblock \href {https://doi.org/10.18653/v1/W19-3805} {Evaluating the underlying gender bias in contextualized word embeddings}.
\newblock In \emph{Proceedings of the First Workshop on Gender Bias in Natural Language Processing}, pages 33--39, Florence, Italy. Association for Computational Linguistics.

\bibitem[{Bender and Friedman(2018)}]{bias-annotation3}
Emily~M. Bender and Batya Friedman. 2018.
\newblock \href {https://doi.org/10.1162/tacl_a_00041} {Data statements for natural language processing: Toward mitigating system bias and enabling better science}.
\newblock \emph{Transactions of the Association for Computational Linguistics}, 6:587--604.

\bibitem[{Bender et~al.(2021)Bender, Gebru, McMillan-Major, and Shmitchell}]{stochastic-parrot}
Emily~M. Bender, Timnit Gebru, Angelina McMillan-Major, and Shmargaret Shmitchell. 2021.
\newblock \href {https://doi.org/10.1145/3442188.3445922} {On the dangers of stochastic parrots: Can language models be too big?}
\newblock In \emph{Proceedings of the 2021 ACM Conference on Fairness, Accountability, and Transparency}, FAccT '21, page 610–623, New York, NY, USA. Association for Computing Machinery.

\bibitem[{Bolukbasi et~al.(2016{\natexlab{a}})Bolukbasi, Chang, Zou, Saligrama, and Kalai}]{cosine-bias}
Tolga Bolukbasi, Kai-Wei Chang, James Zou, Venkatesh Saligrama, and Adam Kalai. 2016{\natexlab{a}}.
\newblock Man is to computer programmer as woman is to homemaker? debiasing word embeddings.
\newblock NIPS'16, page 4356–4364, Red Hook, NY, USA. Curran Associates Inc.

\bibitem[{Bolukbasi et~al.(2016{\natexlab{b}})Bolukbasi, Chang, Zou, Saligrama, and Kalai}]{bias-data1}
Tolga Bolukbasi, Kai-Wei Chang, James~Y. Zou, Venkatesh Saligrama, and Adam~Tauman Kalai. 2016{\natexlab{b}}.
\newblock \href {https://api.semanticscholar.org/CorpusID:1704893} {Man is to computer programmer as woman is to homemaker? debiasing word embeddings}.
\newblock In \emph{Neural Information Processing Systems}.

\bibitem[{Buolamwini and Gebru(2018)}]{bias-annotation2}
Joy Buolamwini and Timnit Gebru. 2018.
\newblock \href {https://proceedings.mlr.press/v81/buolamwini18a.html} {Gender shades: Intersectional accuracy disparities in commercial gender classification}.
\newblock In \emph{Proceedings of the 1st Conference on Fairness, Accountability and Transparency}, volume~81 of \emph{Proceedings of Machine Learning Research}, pages 77--91. PMLR.

\bibitem[{Caliskan et~al.(2017{\natexlab{a}})Caliskan, Bryson, and Narayanan}]{bias-data2}
Aylin Caliskan, Joanna Bryson, and Arvind Narayanan. 2017{\natexlab{a}}.
\newblock Semantics derived automatically from language corpora contain human-like biases.
\newblock \emph{Science}, 356:183--186.

\bibitem[{Caliskan et~al.(2017{\natexlab{b}})Caliskan, Bryson, and Narayanan}]{weat}
Aylin Caliskan, Joanna~J. Bryson, and Arvind Narayanan. 2017{\natexlab{b}}.
\newblock \href {https://doi.org/10.1126/science.aal4230} {Semantics derived automatically from language corpora contain human-like biases}.
\newblock \emph{Science}, 356(6334):183--186.

\bibitem[{Cheng et~al.(2021)Cheng, Hao, Yuan, Si, and Carin}]{debias-sentence-level-2}
Pengyu Cheng, Weituo Hao, Siyang Yuan, Shijing Si, and Lawrence Carin. 2021.
\newblock Fairfil: Contrastive neural debiasing method for pretrained text encoders.
\newblock \emph{arXiv preprint arXiv:2103.06413}.

\bibitem[{Dhamala et~al.(2021{\natexlab{a}})Dhamala, Sun, Kumar, Krishna, Pruksachatkun, Chang, and Gupta}]{bold}
Jwala Dhamala, Tony Sun, Varun Kumar, Satyapriya Krishna, Yada Pruksachatkun, Kai-Wei Chang, and Rahul Gupta. 2021{\natexlab{a}}.
\newblock Bold: Dataset and metrics for measuring biases in open-ended language generation.
\newblock In \emph{Proceedings of the 2021 ACM conference on fairness, accountability, and transparency}, pages 862--872.

\bibitem[{Dhamala et~al.(2021{\natexlab{b}})Dhamala, Sun, Kumar, Krishna, Pruksachatkun, Chang, and Gupta}]{BOLD-debias}
Jwala Dhamala, Tony Sun, Varun Kumar, Satyapriya Krishna, Yada Pruksachatkun, Kai-Wei Chang, and Rahul Gupta. 2021{\natexlab{b}}.
\newblock \href {https://doi.org/10.1145/3442188.3445924} {Bold: Dataset and metrics for measuring biases in open-ended language generation}.
\newblock FAccT '21, New York, NY, USA. Association for Computing Machinery.

\bibitem[{Garimella et~al.(2021)Garimella, Amarnath, Kumar, Yalla, Anandhavelu, Chhaya, and Srinivasan}]{debias-training}
Aparna Garimella, Akhash Amarnath, Kiran Kumar, Akash~Pramod Yalla, N~Anandhavelu, Niyati Chhaya, and Balaji~Vasan Srinivasan. 2021.
\newblock He is very intelligent, she is very beautiful? on mitigating social biases in language modelling and generation.
\newblock In \emph{Findings of the Association for Computational Linguistics: ACL-IJCNLP 2021}, pages 4534--4545.

\bibitem[{Guo et~al.(2022)Guo, Yang, and Abbasi}]{auto-debias}
Yue Guo, Yi~Yang, and Ahmed Abbasi. 2022.
\newblock Auto-debias: Debiasing masked language models with automated biased prompts.
\newblock In \emph{Proceedings of the 60th Annual Meeting of the Association for Computational Linguistics (Volume 1: Long Papers)}, pages 1012--1023.

\bibitem[{Hovy and Prabhumoye(2021)}]{bias-algo2}
Dirk Hovy and Shrimai Prabhumoye. 2021.
\newblock \href {https://doi.org/10.1111/lnc3.12432} {Five sources of bias in natural language processing}.
\newblock \emph{Language and Linguistics Compass}, 15.

\bibitem[{Liu et~al.(2019)Liu, Dacon, Fan, Liu, Liu, and Tang}]{debias-sentence-level-1}
Haochen Liu, Jamell Dacon, Wenqi Fan, Hui Liu, Zitao Liu, and Jiliang Tang. 2019.
\newblock Does gender matter? towards fairness in dialogue systems.
\newblock \emph{arXiv preprint arXiv:1910.10486}.

\bibitem[{Maudslay et~al.(2019{\natexlab{a}})Maudslay, Gonen, Cotterell, and Teufel}]{cda}
Rowan~Hall Maudslay, Hila Gonen, Ryan Cotterell, and Simone Teufel. 2019{\natexlab{a}}.
\newblock \href {https://doi.org/10.18653/v1/D19-1530} {It{'}s all in the name: Mitigating gender bias with name-based counterfactual data substitution}.
\newblock In \emph{Proceedings of the 2019 Conference on Empirical Methods in Natural Language Processing and the 9th International Joint Conference on Natural Language Processing (EMNLP-IJCNLP)}, pages 5267--5275, Hong Kong, China. Association for Computational Linguistics.

\bibitem[{Maudslay et~al.(2019{\natexlab{b}})Maudslay, Gonen, Cotterell, and Teufel}]{advanced-cda}
Rowan~Hall Maudslay, Hila Gonen, Ryan Cotterell, and Simone Teufel. 2019{\natexlab{b}}.
\newblock \href {https://doi.org/10.18653/v1/D19-1530} {It{'}s all in the name: Mitigating gender bias with name-based counterfactual data substitution}.
\newblock In \emph{Proceedings of the 2019 Conference on Empirical Methods in Natural Language Processing and the 9th International Joint Conference on Natural Language Processing (EMNLP-IJCNLP)}, pages 5267--5275, Hong Kong, China. Association for Computational Linguistics.

\bibitem[{May et~al.(2019)May, Wang, Bordia, Bowman, and Rudinger}]{seat}
Chandler May, Alex Wang, Shikha Bordia, Samuel Bowman, and Rachel Rudinger. 2019.
\newblock \href {https://doi.org/10.18653/v1/N19-1063} {On measuring social biases in sentence encoders}.
\newblock pages 622--628.

\bibitem[{Nadeem et~al.(2021)Nadeem, Bethke, and Reddy}]{stereoset}
Moin Nadeem, Anna Bethke, and Siva Reddy. 2021.
\newblock \href {https://doi.org/10.18653/v1/2021.acl-long.416} {{S}tereo{S}et: Measuring stereotypical bias in pretrained language models}.
\newblock In \emph{Proceedings of the 59th Annual Meeting of the Association for Computational Linguistics and the 11th International Joint Conference on Natural Language Processing (Volume 1: Long Papers)}, pages 5356--5371, Online. Association for Computational Linguistics.

\bibitem[{Otterbacher et~al.(2018)Otterbacher, Checco, Demartini, and Clough}]{bias-annotation1}
Jahna Otterbacher, Alessandro Checco, Gianluca Demartini, and Paul Clough. 2018.
\newblock \href {https://doi.org/10.1145/3209978.3210094} {Investigating user perception of gender bias in image search: The role of sexism}.
\newblock In \emph{The 41st International ACM SIGIR Conference on Research \& Development in Information Retrieval}, SIGIR '18, page 933–936, New York, NY, USA. Association for Computing Machinery.

\bibitem[{Perwass(2009)}]{vector-rejection}
Christian Perwass. 2009.
\newblock \emph{Geometric Algebra with Applications in Engineering}, 1st edition.
\newblock Springer Publishing Company, Incorporated.

\bibitem[{Qian et~al.(2022)Qian, Ross, Fernandes, Smith, Kiela, and Williams}]{panda}
Rebecca Qian, Candace Ross, Jude Fernandes, Eric~Michael Smith, Douwe Kiela, and Adina Williams. 2022.
\newblock \href {https://doi.org/10.18653/v1/2022.emnlp-main.646} {Perturbation augmentation for fairer {NLP}}.
\newblock In \emph{Proceedings of the 2022 Conference on Empirical Methods in Natural Language Processing}, pages 9496--9521, Abu Dhabi, United Arab Emirates. Association for Computational Linguistics.

\bibitem[{Radford et~al.(2019)Radford, Wu, Child, Luan, Amodei, and Sutskever}]{zero-shot-task-descripton}
Alec Radford, Jeff Wu, Rewon Child, David Luan, Dario Amodei, and Ilya Sutskever. 2019.
\newblock \href {https://api.semanticscholar.org/CorpusID:160025533} {Language models are unsupervised multitask learners}.

\bibitem[{Ravfogel et~al.(2020)Ravfogel, Elazar, Gonen, Twiton, and Goldberg}]{debias-word-level}
Shauli Ravfogel, Yanai Elazar, Hila Gonen, Michael Twiton, and Yoav Goldberg. 2020.
\newblock Null it out: Guarding protected attributes by iterative nullspace projection.
\newblock \emph{arXiv preprint arXiv:2004.07667}.

\bibitem[{Schick et~al.(2021)Schick, Udupa, and Schütze}]{self-debias}
Timo Schick, Sahana Udupa, and Hinrich Schütze. 2021.
\newblock \href {https://doi.org/10.1162/tacl_a_00434} {{Self-Diagnosis and Self-Debiasing: A Proposal for Reducing Corpus-Based Bias in NLP}}.
\newblock \emph{Transactions of the Association for Computational Linguistics}, 9:1408--1424.

\bibitem[{Sheng et~al.(2019)Sheng, Chang, Natarajan, and Peng}]{regard}
Emily Sheng, Kai-Wei Chang, Premkumar Natarajan, and Nanyun Peng. 2019.
\newblock The woman worked as a babysitter: On biases in language generation.
\newblock \emph{arXiv preprint arXiv:1909.01326}.

\bibitem[{Solaiman et~al.(2019)Solaiman, Brundage, Clark, Askell, Herbert-Voss, Wu, Radford, Krueger, Kim, Kreps et~al.}]{bias-algo1}
Irene Solaiman, Miles Brundage, Jack Clark, Amanda Askell, Ariel Herbert-Voss, Jeff Wu, Alec Radford, Gretchen Krueger, Jong~Wook Kim, Sarah Kreps, et~al. 2019.
\newblock Release strategies and the social impacts of language models.
\newblock \emph{arXiv preprint arXiv:1908.09203}.

\bibitem[{Su et~al.(2023)Su, Shi, Kasai, Wang, Hu, Ostendorf, Yih, Smith, Zettlemoyer, and Yu}]{Instructor}
Hongjin Su, Weijia Shi, Jungo Kasai, Yizhong Wang, Yushi Hu, Mari Ostendorf, Wen-tau Yih, Noah~A. Smith, Luke Zettlemoyer, and Tao Yu. 2023.
\newblock \href {https://doi.org/10.18653/v1/2023.findings-acl.71} {One embedder, any task: Instruction-finetuned text embeddings}.
\newblock In \emph{Findings of the Association for Computational Linguistics: ACL 2023}, pages 1102--1121, Toronto, Canada. Association for Computational Linguistics.

\bibitem[{Zhang et~al.(2020)Zhang, Sun, Galley, Chen, Brockett, Gao, Gao, Liu, and Dolan}]{DIALOGPT}
Yizhe Zhang, Siqi Sun, Michel Galley, Yen-Chun Chen, Chris Brockett, Xiang Gao, Jianfeng Gao, Jingjing Liu, and Bill Dolan. 2020.
\newblock \href {https://doi.org/10.18653/v1/2020.acl-demos.30} {{DIALOGPT} : Large-scale generative pre-training for conversational response generation}.
\newblock In \emph{Proceedings of the 58th Annual Meeting of the Association for Computational Linguistics: System Demonstrations}, pages 270--278, Online. Association for Computational Linguistics.

\bibitem[{Zhao et~al.(2019)Zhao, Wang, Yatskar, Cotterell, Ordonez, and Chang}]{word-level-bias}
Jieyu Zhao, Tianlu Wang, Mark Yatskar, Ryan Cotterell, Vicente Ordonez, and Kai-Wei Chang. 2019.
\newblock \href {https://doi.org/10.18653/v1/N19-1064} {Gender bias in contextualized word embeddings}.
\newblock In \emph{Proceedings of the 2019 Conference of the North {A}merican Chapter of the Association for Computational Linguistics: Human Language Technologies, Volume 1 (Long and Short Papers)}, pages 629--634, Minneapolis, Minnesota. Association for Computational Linguistics.

\bibitem[{Zhao et~al.(2018)Zhao, Wang, Yatskar, Ordonez, and Chang}]{wino}
Jieyu Zhao, Tianlu Wang, Mark Yatskar, Vicente Ordonez, and Kai-Wei Chang. 2018.
\newblock \href {https://doi.org/10.18653/v1/N18-2003} {Gender bias in coreference resolution: Evaluation and debiasing methods}.
\newblock In \emph{Proceedings of the 2018 Conference of the North {A}merican Chapter of the Association for Computational Linguistics: Human Language Technologies, Volume 2 (Short Papers)}, pages 15--20, New Orleans, Louisiana. Association for Computational Linguistics.

\bibitem[{Zheng et~al.(2023)Zheng, Zhou, Meng, Zhou, and Huang}]{debias-MCQ}
Chujie Zheng, Hao Zhou, Fandong Meng, Jie Zhou, and Minlie Huang. 2023.
\newblock On large language models' selection bias in multi-choice questions.
\newblock \emph{arXiv preprint arXiv:2309.03882}.

\bibitem[{Zhu et~al.(2023)Zhu, Liu, Dong, Xu, Kong, Chen, Li, and Huang}]{translate-LLM}
Wenhao Zhu, Hongyi Liu, Qingxiu Dong, Jingjing Xu, Lingpeng Kong, Jiajun Chen, Lei Li, and Shujian Huang. 2023.
\newblock Multilingual machine translation with large language models: Empirical results and analysis.
\newblock \emph{arXiv preprint arXiv:2304.04675}.

\end{thebibliography}
\bibliographystyle{acl_natbib}

\appendix

\end{document}